\documentclass[aps,pre,reprint,amsmath]{revtex4-2}

\usepackage[pdftex]{graphicx}
\usepackage{newtx}
\usepackage{here}
\usepackage{url}
\usepackage{bm}
\usepackage{algorithm}
\usepackage{algpseudocode}
\usepackage{stmaryrd}

\usepackage{blkarray}

\usepackage{hyperref}

\newcommand{\tsor}[1]{\mathbf{#1}}
\newcommand{\ts}{\mathsf{T}}
\newcommand{\argmin}{\mathop{\rm arg\,min}\limits}
\newcommand{\matvec}[2]{\begin{bmatrix} #1 & \cdots & #2 \end{bmatrix}}
\newcommand{\matbb}[1]{\left\llbracket #1 \right\rrbracket}

\newcommand{\ndim}{d}
\newcommand{\ndic}{N_{\mathrm{dic}}}
\newcommand{\ndata}{N_{\mathrm{data}}}
\newcommand{\stime}{T_{\mathrm{s}}}

\let\Im\relax
\DeclareMathOperator{\Im}{Im}

\begin{document}

\title{Tensor-based computation of the Koopman generator via operator logarithm}

\author{Tatsuya Kishimoto and Jun Ohkubo}

\affiliation{Graduate School of Science and Engineering, Saitama University\\
255 Shimo-Okubo, Sakura-ku, Saitama 338--8570, Japan}

\begin{abstract}
Identifying governing equations of nonlinear dynamical systems from data is challenging.
While sparse identification of nonlinear dynamics (SINDy) and its extensions are widely used for system identification, operator-logarithm approaches use the logarithm to avoid time differentiation, enabling larger sampling intervals.
However, they still suffer from the curse of dimensionality.
Then, we propose a data-driven method to compute the Koopman generator in a low-rank tensor train (TT) format by taking logarithms of Koopman eigenvalues while preserving the TT format.
Experiments on 4-dimensional Lotka--Volterra and 10-dimensional Lorenz--96 systems show accurate recovery of vector field coefficients and scalability to higher-dimensional systems.
\end{abstract}

\maketitle

\section{Introduction}
\label{introduction}
The identification of governing equations for nonlinear dynamical systems from data has been actively studied in many fields, such as control theory, physics, and biology.
This task is frequently challenging due to the inherent nonlinearity of the underlying systems.
In the context of nonlinear parameter estimation, various methods have been developed to identify the state dynamics of autonomous systems; sparse identification of nonlinear dynamics (SINDy) \cite{brunton_2016_discovering} has attracted particular attention among these methods.
SINDy identifies the vector field of deterministic dynamical systems by solving a sparse regression problem using a predefined dictionary of candidate functions.
Several extensions of this framework have already been carried out \cite{boninsegna_2018_sparse,huang_2022_sparse}.

In this paper, we focus on the method based on the Koopman operator \cite{koopman_1931_hamiltonian} framework, which have been applied in various research areas including molecular dynamics, meteorology, and economics.
By lifting nonlinear dynamics to an infinite-dimensional linear operator acting on observables, this framework facilitates the analysis of nonlinear systems using linear techniques.
However, since the Koopman operator is infinite-dimensional, it cannot be handled directly in numerical computations and necessitates approximation within a finite-dimensional space.

To this end, several data-driven algorithms have been proposed.
Dynamic mode decomposition (DMD) \cite{rowley_2009_spectral,schmid_2010_dynamic} extracts spectral information of the Koopman operator from time-series data, and its variants have been applied to partial differential equations such as the Burgers equation.
Extended dynamic mode decomposition (EDMD) \cite{williams_2015_data} generalizes DMD by employing a dictionary of basis functions to obtain a finite-dimensional approximation.
Applications of EDMD are extensive, covering forecasting, system identification, and control; see \cite{mezic_2021_koopman,brunton_2022_modern} for more details.

Recently, Koopman-based methods have been specifically adapted for system identification.
The Koopman generator, which is the infinitesimal generator of the Koopman semigroup, describes the time derivative of observables.
An extension of EDMD, known as generator EDMD (gEDMD) \cite{klus_2020_datadriven}, can estimate ordinary differential equations or stochastic differential equations from data.
Direct system identification methods, such as SINDy and gEDMD, aim to identify the governing vector field directly from data and typically require estimating time derivatives from data.
This estimation is often sensitive to measurement noise and the choice of sampling interval.

In contrast, indirect methods that solve an initial value problem avoid explicit derivative estimation, thereby allowing for longer sampling intervals.
While solving an initial value problem typically involves nonlinear least-squares optimization, an operator-logarithm-based method \cite{mauroy_2020_koopmanbased} offers an alternative by identifying the system via a linear least-squares problem.
This approach involves computing a finite-dimensional Koopman operator using EDMD and subsequently taking its logarithm to approximate the Koopman generator.

However, for high-dimensional systems, these methods suffer from the curse of dimensionality, resulting in computational costs and memory requirements that grow exponentially with dimensionality.
As the number of entries in a tensor grows exponentially with dimensionality, low-rank tensor formats requiring only a tractable number of parameters become essential.
Prominent examples include the canonical format \cite{carroll_1970_analysis}, the Tucker format \cite{tucker_1966_mathematical}, and the tensor train (TT) format \cite{oseledets_2009_new,oseledets_2011_tensortrain}, all of which represent high-dimensional tensors as networks of lower-dimensional components.
The TT format is also known as matrix product state (MPS) \cite{perez-garcia_2007_matrix,schollwock_2011_densitymatrix} in quantum physics, and was applied to many fields, such as machine learning \cite{novikov_2015_tensorizing} and the numerical solution of partial differential equations \cite{chertkov_2021_solution}.

Recent studies have leveraged such formats to represent the Koopman operator and its eigenfunctions in high-dimensional settings.
For instance, the AMUSEt algorithm \cite{nuske_2021_tensorbased} efficiently computes Koopman eigenfunctions in the TT format, and a method combining the TT format with gEDMD (tgEDMD) \cite{lucke_2022_tgedmd} computes the generator directly in the TT format.
Despite these advances, the operator-logarithm-based method for computing the Koopman generator within the TT format remains undeveloped, primarily because computing a matrix logarithm while preserving the TT structure presents a significant challenge.

We propose a data-driven method to compute the Koopman generator in the TT format by extending the operator-logarithm approach.
Our proposed method introduces a way to take the operator logarithm directly in the TT format by leveraging eigendecomposition.
Specifically, we first compute the Koopman eigenvalues and the corresponding eigenfunctions represented in the TT format.
We then take the logarithm of the eigenvalues and use the resulting eigendecomposition to assemble the Koopman generator in the TT format.
In this way, eigendecomposition serves as the key mechanism that enables a TT-format logarithm without explicitly computing a matrix logarithm or breaking the TT structure.
We demonstrate the effectiveness of the proposed method through numerical experiments on a 4-dimensional Lotka--Volterra system and a 10-dimensional Lorenz--96 system, the latter serving as a benchmark for higher-dimensional dynamics.

This paper is organized as follows.
In Sec.~\ref{sec:background}, we review the Koopman operator, the Koopman generator, and methods for identifying the Koopman generator from data.
In Sec.~\ref{sec:koopman_eig}, we revisit the relationship between the Koopman operator and its generator, particularly focusing on their eigendecomposition.
In Sec.~\ref{sec:proposed}, we present the proposed method to compute the Koopman generator in the TT format in a data-driven manner.
In Sec.~\ref{sec:experiments}, we demonstrate the effectiveness of the proposed method through numerical experiments.
Finally, we conclude the paper in Sec.~\ref{sec:conclusion}.

\section{Background}
\label{sec:background}

  \subsection{Dynamical systems and Koopman theory}
  We consider the continuous-time nonlinear dynamical system
  \begin{equation}
    \dot{\bm{x}} = \bm F(\bm{x}), \quad \bm{x}(t) \in \mathbb{R}^{\ndim},
  \end{equation}
  where $\bm F: \mathbb{R}^{\ndim} \to \mathbb{R}^{\ndim}$ is a nonlinear vector field.
  Although $\bm F$ is nonlinear, the Koopman operator is a linear time-evolution operator acting on a lifted (function) space, whose elements are called observables.
  Let $f: \mathbb{R}^{\ndim}\to\mathbb{C} \in \mathcal{F}$ be an observable, and let $S^t:\mathbb{R}^{\ndim}\to\mathbb{R}^{\ndim}$ denote the flow map generated by $\bm F$ over a time interval  $t \ge 0$.
  That is, if $\bm x(t)$ is the solution of the initial value problem with $\bm x(0)=\bm x_0$, then $S^t(\bm x_0)=\bm x(t)$.
  The Koopman operator $\mathcal{K}_t$ is defined by
  \begin{equation}
    \mathcal{K}_t f = f \circ S^t.
  \end{equation}

  By the properties of the flow map, it is easy to verify that $\{ \mathcal{K}_t \}_{t \geq 0}$ forms a semigroup.
  The Koopman semigroup is strongly continuous and generated by the infinitesimal generator
  \begin{equation}
  \label{eq:generator_def}
    \mathcal{L}f \coloneq \lim_{t \to 0} \frac{\mathcal{K}_t f - f}{t}.
  \end{equation}
  If the observable function are differentiable, we can verify that the Koopman generator is such that
  \begin{equation}
    \label{eq:koopman_generator_impl}
    \frac{\partial (f \circ S^t)}{\partial t} = \nabla f \cdot \bm F = \mathcal{L}f.
  \end{equation}
  Here, $\partial (f \circ S^t)/\partial t$ is the ``time derivative'' of the observable function $f$ along the flow map $S^t$, and we denote it by $\partial f /\partial t$ for brevity.
  If $\mathcal{L}$ is a bounded operator,
  \begin{equation}
  \label{eq:op_exp}
    \mathcal{K}_t = e^{t \mathcal{L}}
  \end{equation}
  for each $t \geq 0$.

  \subsection{Identification of the Koopman generator matrix}
  \label{ssec:koopman_generator_identification}
  Since the Koopman operator and generator are infinite-dimensional operators, we need to approximate them in a finite-dimensional space.
  For this purpose, we consider approximating the infinite-dimensional space $\mathcal{F}$ by a finite-dimensional subspace $\mathcal{F}_N \subset \mathcal{F}$ spanned by a set of basis functions $\{\psi_1, \dots, \psi_{\ndic}\}$, where $\ndic$ is the size of the dictionary.
  The choice of basis functions is problem-dependent and should be selected based on the characteristics of the system being studied.

  In this paper, we define the dictionary as
  \begin{equation}
    \bm{\psi}(\bm{x}) = \matvec{\psi_1(\bm{x})}{\psi_{\ndic}(\bm{x})}^\ts,
  \end{equation}
  where $\psi_i$ is the $i$-th basis function.
  Note that in \cite{williams_2015_data}, the dictionary is defined as a row vector, but in this paper, we consider it as a column vector.
  Given $\ndata$ snapshot pairs $\{(\bm{x}_k, \bm{y}_k)\}_{k=1}^{\ndata}$, where $\bm{y}_k = S^{\stime}(\bm{x}_k)$ and $\stime$ is the sampling interval, the data matrices are defined as $X = [\bm{x}_1, \dots, \bm{x}_{\ndata}]$ and $Y = [\bm{y}_1, \dots, \bm{y}_{\ndata}]$.
  The finite-dimensional approximation of the Koopman operator $K \in \mathbb{R}^{\ndic \times \ndic}$ is then defined as the solution to
  \begin{equation}
    K = \argmin_{\widetilde{K}} \sum_{i=1}^{\ndata} \| \bm{\psi}(\bm y_i) - \widetilde{K} \bm{\psi}(\bm x_i) \|^2.
  \end{equation}

  To derive a finite-dimensional approximation of the Koopman generator $L \in \mathbb{R}^{\ndic \times \ndic}$ from $K$, we use the operator logarithm shown in Eq.~\eqref{eq:op_exp}, i.e.,
  \begin{equation}
    L = \frac{1}{\stime}\log(K).
  \end{equation}
  As Eq.~\eqref{eq:koopman_generator_impl} shows, the Koopman generator describes the time derivative of the observable function.
  Since $L$ is the finite-dimensional approximation of the Koopman generator, it is possible to compute the time derivative of the basis function in the finite-dimensional space as
  \begin{equation}
    \frac{\partial \psi_j}{\partial t} = \sum_{i=1}^{\ndic} L_{ji} \psi_i.
  \end{equation}

  \subsection{System identification using the Koopman generator}
  We consider the vector field $\bm F(\bm{x})$ is of the form
  \begin{equation}
    F_k(\bm{x}) = \bm{w}_k^{\ts} \bm{\psi}(\bm x),
  \end{equation}
  where $\bm{\psi}(\bm x)$ is the dictionary and $\bm w_k^{\ts} = [w_{k,1}, \dots, w_{k,\ndic}]$ is the coefficient vector for the $k$-th dimension.
  In order to identify the vector field coefficients $\bm w_k$, we add the full-state observable $g_k(\bm x) = x_k$, $k=1,\dots,d$ to the dictionary.
  From the definition of the Koopman generator in Eq.~\eqref{eq:koopman_generator_impl}, we have
  \begin{equation}
    \mathcal{L} g_k = \sum_{i=1}^{\ndic} F_i(\bm{x})\frac{\partial g_k}{\partial x_i} = F_k(\bm{x})
  \end{equation}
  in the finite-dimensional setting.
  Therefore, if we have the Koopman generator in a finite-dimensional space that includes the full-state observable, we can identify the vector field coefficients by
  \begin{equation}
    F_k = \sum_{i=1}^{\ndic} L_{ki} \psi_i,
  \end{equation}
  i.e., the $k$-th row of the $L$ represents the vector field coefficients $\bm{w}_k^{\ts}$.

  \subsection{Example of system identification using the Koopman generator}
\begin{figure}[t]
    \centering
    \includegraphics[keepaspectratio, scale=0.6]{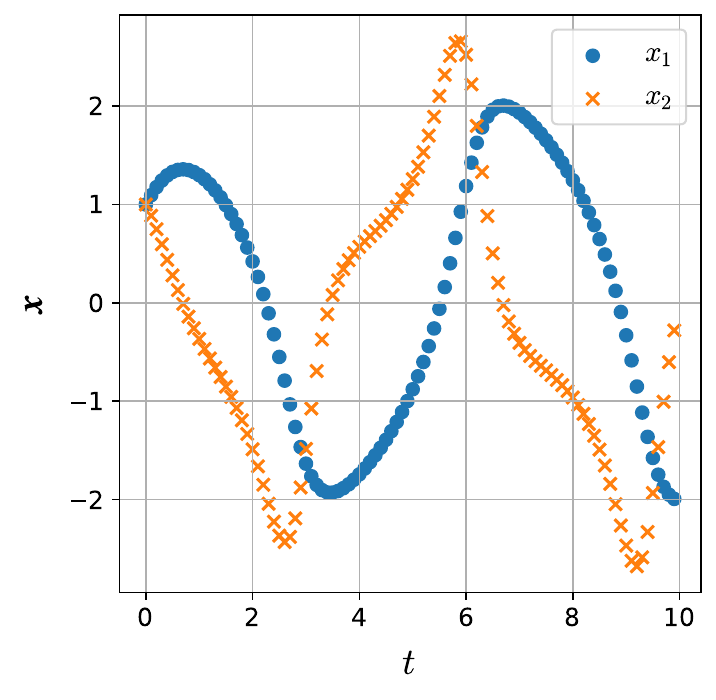}
    \caption{Segment of the vdP oscillator trajectory data used for EDMD.
The trajectory is generated from the initial condition $[x_1,x_2]=[1.0,1.0]$, and this plot shows the part from $t = 0$ to $t = 10$ sampled with a sampling interval of $\stime=0.1$.}
    \label{fig:vdp}
  \end{figure}

  As a simple example of system identification via the Koopman generator, we consider the van der Pol (vdP) oscillator
  \begin{equation}
    \begin{aligned}
      \dot{x}_1 &= x_2,\\
      \dot{x}_2 &= (1-x_1^2)x_2 - x_1.
    \end{aligned}
  \end{equation}
  We assume that the vector field can be represented by a polynomial.
  Accordingly, we construct the dictionary of monomials up to degree two in each variable:
  \begin{equation}
    \bm{\psi}(\bm{x}) =
    \begin{bmatrix}
      1 & x_1 & x_1^2 & x_2 & x_1x_2 & x_1^2x_2 & \dots & x_1^2x_2^2
    \end{bmatrix}^{\ts}.
  \end{equation}
  We generate a single trajectory starting from the initial condition $[x_1,x_2]=[1.0,1.0]$ using \verb|scipy.integrate.solve_ivp| in Python with the explicit Runge--Kutta method of order 5(4) (RK45).
  We then sample 1000 snapshot pairs at a sampling interval of $\stime=0.1$.
  Figure~\ref{fig:vdp} shows a segment of the resulting trajectory consisting of the first 100 data points, i.e., those from $t=0$ to $t=10$.
  We apply EDMD to compute the Koopman matrix and then compute the generator via the matrix logarithm.
  To take the matrix logarithm, we employ \verb|scipy.linalg.logm| in Python.

  The transpose of the estimated generator $L^{\ts}$ is evaluated as follows:
  \begin{equation}
  \label{eq:vdp_generator}
      L^{\ts} =
      \begin{blockarray}{c*{5}{c}}
      & 1 & dx_1 & dx_1^2 & dx_2 & \cdots \\
      \begin{block}{c[*{5}{c}]}
      1        & 0.00 &  0.00 & 0.01 &  0.00 & \\
      x_1      & 0.00 &  0.00 & 0.00 & -1.00 & \\
      x_1^2    & 0.00 &  0.00 & 0.00 &  0.00 & \\
      x_2      & 0.00 &  1.00 & 0.00 &  1.00 & \cdots \\
      x_1x_2   & 0.00 &  0.00 & 2.00 &  0.00 & \\
      x_1^2x_2 & 0.00 &  0.00 & 0.00 & -1.00 & \\
      \vdots & & & \vdots \\
      \end{block}
      \end{blockarray},
  \end{equation}
  where the rows correspond to the basis functions ordered as $1, x_1, x_1^2, x_2, \dots$, and the columns correspond to their time derivatives in the same order.
  For readability, the matrix entries are rounded to two decimal places.
  The estimated coefficients in the second and fourth columns match the true coefficients.

\section{Revisit on Koopman generator estimation}
\label{sec:koopman_eig}
In this section, we revisit the relationship between the Koopman operator and its generator, focusing in particular on eigendecompositions and the matrix logarithm of the Koopman operator.

  \subsection{Koopman eigenvalues and eigenfunctions}
  Consider the Koopman operator $\mathcal{K}_t$ associated with the flow map $S^t$ and its generator $\mathcal{L}$.
  When both operators are bounded, their spectra $\sigma(\cdot)$ satisfy
  \begin{equation}
  \label{eq:spectrum_exp}
    \sigma(\mathcal{K}_t) = e^{t \sigma \left(\mathcal{L}\right)} = \{e^{\lambda t} \mid \lambda \in \sigma(\mathcal{L})\},
  \end{equation}
  for all $t \geq 0$.
  An eigenfunction $\phi \in \mathcal{F} \setminus \{0\}$ of $\mathcal{K}_t$ and its corresponding eigenvalue $\mu \in \mathbb{C}$ satisfy the following relation:
  \begin{equation}
    \label{eq:op_eigen}
    \mathcal{K}_t \phi = \mu \phi = e^{\lambda t}\phi, \quad \forall t \geq 0.
  \end{equation}
  Here, $\lambda$ is the eigenvalue of $\mathcal{L}$.
  Assuming Eq.~\eqref{eq:generator_def} defines the Koopman generator, $\lambda$ and $\phi$ are also an eigenvalue-eigenfunction pair for the Koopman generator $\mathcal{L}$:
  \begin{equation}
    \label{eq:generator_eigen}
    \mathcal{L} \phi = \lambda \phi.
  \end{equation}
  We call $\phi$ a Koopman eigenfunction and $\lambda$ the associated Koopman eigenvalue.
  The key points are that $\mathcal{K}_t$ and $\mathcal{L}$ share the same eigenfunction $\phi$, and that their respective eigenvalues, $\mu$ and $\lambda$, are related via the exponential map $\mu = e^{\lambda t}$ as shown in Eq.~\eqref{eq:spectrum_exp}.

  \subsection{Matrix logarithm and eigendecomposition}
  We now revisit the relationship between the Koopman operator and its generator in the finite-dimensional setting.
  If the Koopman operator matrix $K$ is eigendecomposed as $K = P \Sigma P^{-1}$, the matrix logarithm is computed as
  \begin{equation}
  \label{eq:matrix_log}
    L = \frac{1}{\stime}\log(P \Sigma P^{-1}) = \frac{1}{\stime} P \log(\Sigma) P^{-1}.
  \end{equation}
  From this equation, we can see that the eigenvectors of $K$ and $L$ are identical, and the eigenvalues of $L$ are derived by taking the logarithm of the eigenvalues of $K$.
  This relationship is consistent with the relationship between the infinite-dimensional operators shown in Eqs.~\eqref{eq:spectrum_exp} and \eqref{eq:op_eigen}.

  When computing the logarithm of the Koopman eigenvalues, we should be careful about which branch of the logarithm to use, because the Koopman eigenvalues are complex-valued.
  The logarithm of a complex number is defined only up to an integer multiple of $2\pi i$:
  \begin{equation}
    \begin{aligned}
      \log(e^z) &= \log\left( e^x \left( \cos y + i \sin y \right) \right)\\
                &= x + i\arg\left( \cos y + i \sin y \right)\\
                &= x + i\left( y + 2\pi k \right), \quad k \in \mathbb{Z},
    \end{aligned}
  \end{equation}
  where $z = x + iy$ is a complex number.
  This branch ambiguity imposes a constraint on the sampling interval \cite{zeng_2024_sampling}.
  If the model equations and dictionary are given, it is possible to compute the true Koopman generator, and we have the eigenvalues of the Koopman generator.
  Then, as shown in \cite{zeng_2024_sampling}, the sampling interval $\stime$ must be chosen to satisfy
  \begin{equation}
    \label{eq:ts_limit}
    \stime < \frac{\pi}{\max |\Im(\sigma(L))|}.
  \end{equation}
  When the governing equations are unknown, we sample trajectories with a sufficiently small time interval.
  We then vary the effective sampling interval using snapshot pairs with different time lags, such as $\left\{(x_i,y_i), (x_i,y_{i+1}), (x_i,y_{i+2}),\dots\right\}$, and verify that the estimated Koopman generator is not affected by the branch ambiguity of the logarithm.

  However, even if the branch ambiguity is properly addressed, the direct computation of the Koopman generator becomes infeasible in high-dimensional systems.
  This is because the size of the Koopman matrix $K$ grows exponentially with the dimension of the original system.
  If we use a dictionary of monomials up to degree $n$ for each dimension when the system is $d$-dimensional, the size of the dictionary would be
  \begin{equation}
    \ndic = (n+1)^d.
  \end{equation}
  For example, if we use monomials up to degree two in each variable for a 10-dimensional system, the size of the dictionary would be $\ndic = 3^{10} = 59,049$.
  In this case, the Koopman matrix $K$ would have $59,049^2 \approx 3.5 \times 10^9$ entries, which is computationally infeasible to store and compute with.
  Therefore, storing and computing the Koopman matrix $K$ and taking its logarithm become infeasible due to memory constraints.

\section{Proposal: identification of the Koopman generator in TT format}
\label{sec:proposed}
This section describes how to construct Koopman generators in higher-dimensional systems.
To obtain a finite-dimensional approximation, we consider tensor product spaces built from elementary subspaces of moderate dimensions.
Accordingly, many quantities, such as the dictionary and the Koopman matrix, naturally admit a tensor structure.

  \subsection{TT format}
  The choice of the finite-dimensional subspace $\mathcal{F}_N$ in Sec.~\ref{ssec:koopman_generator_identification} is critical to the quality of the Koopman operator approximation.
  A common way to obtain a rich approximation space is to take tensor products of functions from a collection of elementary subspaces of moderate dimensions.
  Here, we yield a brief introduction to the TT format.
  For details, see \cite{oseledets_2011_tensortrain}.

  For example, we consider a tensor $\tsor{T}$.
  For our purposes, it is sufficient to view a tensor as a multi-dimensional array whose entries are indexed by $p$ indices:
  \begin{equation}
    \tsor{T}_{i_1,\dots,i_p} \in \mathbb{R}.
  \end{equation}
  The number of indices $p$ is called the \emph{order} of the tensor.
  The indices are often referred to as \emph{modes}, with mode sizes $n_k$.

  The tensor product of $p$ vectors $\bm{v}^{(k)} \in \mathbb{R}^{n_k}$ is defined by
  \begin{equation}
    \tsor{T} = \bigotimes_{k=1}^p \bm{v}^{(k)}, \quad \tsor{T}_{i_1,\dots,i_p} = \prod_{k=1}^p v^{(k)}_{i_k},
  \end{equation}
  which generalizes the outer product of two vectors.
  A tensor of this form is called a rank-one tensor.
  We will also use the tensor product for matrices, known as the Kronecker product, and denote it by the same symbol.
  In particular, for vectors, the Kronecker product corresponds to representing the result of the tensor product as a vector.
  For example, assuming two vectors $\bm a \in \mathbb{R}^m$ and $\bm b \in \mathbb{R}^n$, the Kronecker product yields the following blockwise structures:
  \begin{equation}
    \bm a \otimes \bm b =
    \begin{bmatrix}
      a_1 \bm b\\
      \vdots\\
      a_m \bm b
    \end{bmatrix}
    =
    \begin{bmatrix}
      a_1 b_1\\
      \vdots\\
      a_1 b_n\\
      \vdots\\
      a_m b_1\\
      \vdots\\
      a_m b_n
    \end{bmatrix}
    \in \mathbb{R}^{mn}.
  \end{equation}
  The resulting vector has length $mn$; when taking tensor products across many vectors, this size grows exponentially.

  Every tensor can be expressed as a finite linear combination of rank-one tensors, but the required number of terms may be exponentially large.
  In many applications, tensors can be approximated accurately with few parameters.
  Many such representations, known as tensor formats, have been proposed.
  Here, we focus on the TT format, in which a tensor is represented via contractions of lower-order tensors.
  A tensor $\tsor{T}$ is said to be in the TT format if
  \begin{equation}
    \label{eq:tt_format_elem}
    \tsor{T}_{i_1,\dots,i_d} =
    \sum_{k_0=1}^{r_0}\cdots\sum_{k_d=1}^{r_d}
    \tsor{T}^{(1)}_{k_0,i_1,k_1}\cdots
    \tsor{T}^{(d)}_{k_{d-1},i_d,k_d}.
  \end{equation}
  Equation~\eqref{eq:tt_format_elem} yields the specific $i_1, \dots, i_d$ component of the tensor $\tsor{T}$.
  In the context of the TT format, it is common to use the following notation:
  \begin{equation}
    \tsor{T} =
    \sum_{k_0=1}^{r_0}\cdots\sum_{k_d=1}^{r_d}
    \tsor{T}^{(1)}_{k_0,:,k_1}\otimes\cdots\otimes
    \tsor{T}^{(d)}_{k_{d-1},:,k_d},
  \end{equation}
  which represents the tensor $\tsor{T}$ as a sum of rank-one tensors formed by tensor products of vectors.
  The tensors $\tsor{T}^{(k)} \in \mathbb{R}^{r_{k-1}\times n_k \times r_k}$ of order 3 are called TT cores, and the integers $r_k$ are called TT ranks ($k=0,\dots,d$).
  It holds that $r_0=r_d=1$.
  The TT ranks strongly affect both the representational power of the TT format and the storage cost.

  We also represent TT cores as two-dimensional arrays containing vectors as elements.
  For a given TT format $\tsor{T} \in \mathbb{R}^{n_1 \times \cdots \times n_d}$ with cores $\tsor{T}^{(k)} \in \mathbb{R}^{r_{k-1}\times n_k\times r_k}$, a single core is written as
  \begin{equation}
    \matbb{\tsor{T}^{(k)}} =
    \left\llbracket
    \begin{array}{ccc}
    \tsor{T}^{(k)}_{1,:,1} & \cdots & \tsor{T}^{(k)}_{1,:,r_k} \\
    \vdots & \ddots & \vdots \\
    \tsor{T}^{(k)}_{r_{k-1},:,1} & \cdots & \tsor{T}^{(k)}_{r_{k-1},:,r_k}
    \end{array}
    \right\rrbracket.
  \end{equation}
  We then use the notation $\tsor{T} = \matbb{\tsor{T}^{(1)}} \otimes \cdots \otimes \matbb{\tsor{T}^{(d)}}$ for representing the TT format $\tsor{T}$.
  This notation can be regarded as a generalization of standard matrix multiplication.
  The difference is that, instead of multiplying corresponding scalar entries, we take tensor products of the corresponding vector entries and then sum over the column and row indices, respectively.

  In this paper, we write $\tsor{T}\tsor{U}$ to denote the contraction of two tensors over the last mode of $\tsor{T}$ and the first mode of $\tsor{U}$, i.e.,
  \begin{equation}
    \tsor{T}\tsor{U}=\sum_{k=1}^{r} \tsor{T}_{\cdots,k}\otimes \tsor{U}_{k,\cdots}.
  \end{equation}

  \begin{figure}[t]
    \centering
    \includegraphics[keepaspectratio, scale=0.54]{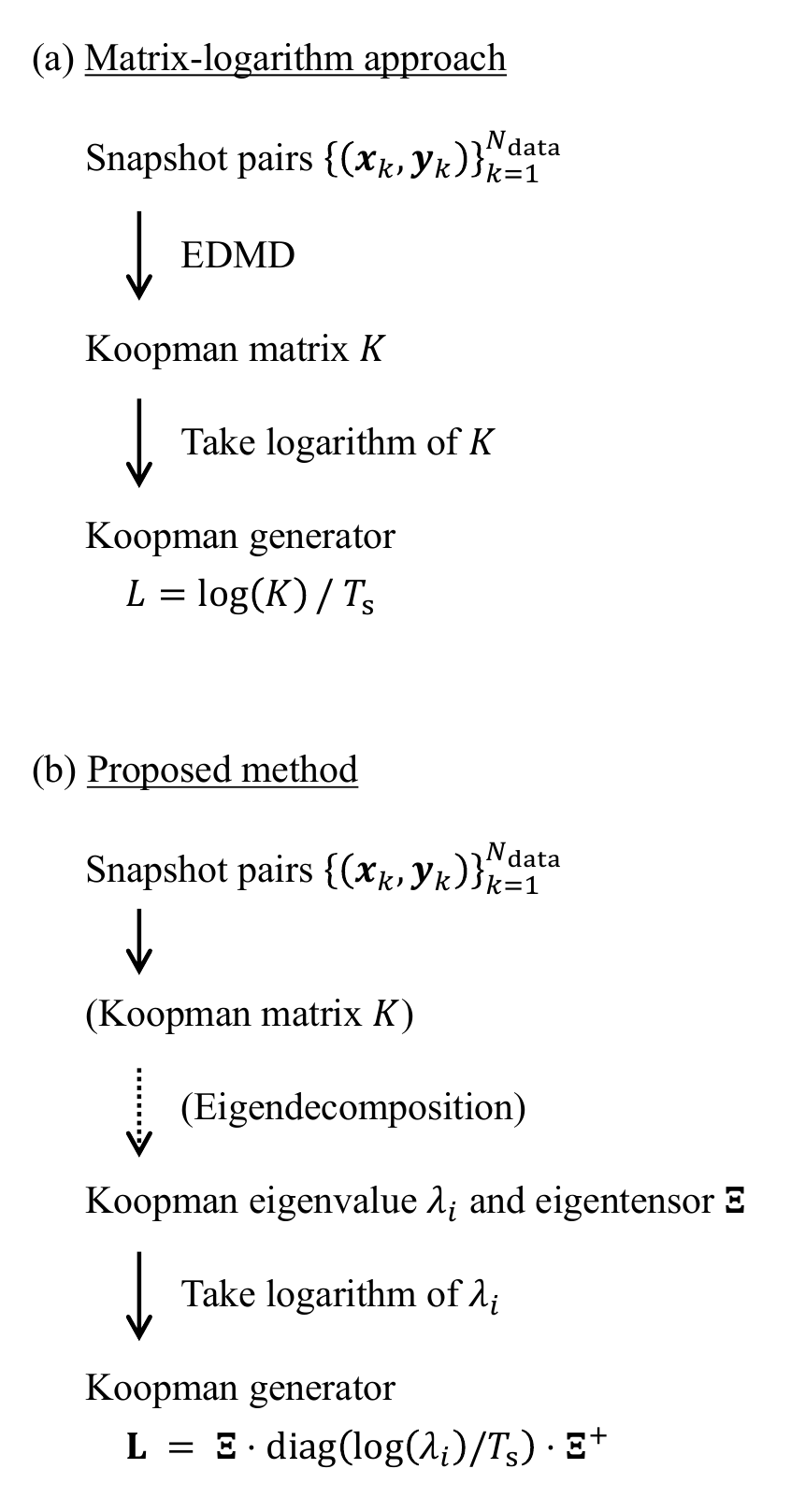}
    \caption{Illustration of the proposed method for computing the Koopman generator in the TT format.
(a) shows the conventional matrix-logarithm-based approach, whereas (b) summarizes the proposed approach based on the eigendecomposition relationship.
In the proposed method, the logarithm is applied to the eigenvalues rather than to the Koopman matrix, enabling efficient computation in high-dimensional settings.
The Koopman matrix and its eigendecomposition are shown in parentheses to indicate that the eigenvalues and eigentensors can be computed directly using AMUSEt.}
    \label{fig:proposed_rep}
  \end{figure}

  \subsection{Direct evaluation of the Koopman generator matrix in TT format}
  Computing the operator logarithm directly in the TT format is difficult because no method has been known for computing the matrix logarithm while preserving the TT structure.
  Therefore, we propose a method to compute the Koopman generator in the TT format by leveraging the eigendecomposition relationship in Eq.~\eqref{eq:matrix_log}, which involves the eigenvalues and eigenvectors.
  In Eq.~\eqref{eq:matrix_log}, if we have $P$ in the TT format, then we can obtain the Koopman generator in the TT format.
  Figure~\ref{fig:proposed_rep} illustrates a comparison between the existing matrix-logarithm-based method and the proposed method.

  First, we define the dictionary as a tensor product of one-dimensional basis functions, i.e.,
  \begin{equation}
    \tsor{\psi}(\bm{x}) = \bigotimes_{i=1}^{\ndim} \bm{\psi}_i(x_i),
  \end{equation}
  where $\bm{\psi}_i(x_i) = \left[ \psi_{i,1}(x_i), \dots, \psi_{i,n}(x_i) \right]$ is the one-dimensional dictionary for the $i$-th dimension.
  With this definition, we can represent $K$ and $L$ as tensors.

\begin{figure}[H]
    \centering
    \includegraphics[keepaspectratio, scale=0.5]{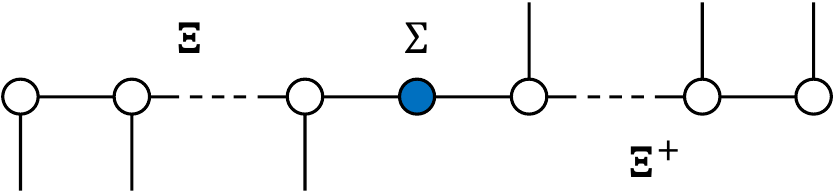}
    \caption{Tensor contraction process for computing the Koopman generator in the TT format.
$\tsor{\Xi}$ is the Koopman eigentensor, $\Sigma$ is the diagonal matrix of the eigenvalues of the Koopman generator, and $\tsor{\Xi}^{+}$ is the pseudo-inverse of the Koopman eigentensor.
By contracting the last mode of $\tsor{\Sigma}$ and the first mode of $\tsor{\Xi}^{+}$, we can obtain the Koopman generator $\tsor{L}$ in the TT format.}
    \label{fig:proposed}
  \end{figure}

  Next, we compute the Koopman eigenvalues $\lambda_i$ and the matrix $P$ of eigenvectors of the Koopman matrix $K$ in the TT format by using the AMUSEt algorithm \cite{nuske_2021_tensorbased}; in Appendix~\ref{sec:appendix}, we provide a brief explanation of the AMUSEt algorithm.
  In the following, we refer to the TT format of $P$ as the Koopman eigentensor $\tsor{\Xi}$.
  The Koopman generator in the TT format can be computed as
  \begin{equation}
  \label{eq:koopman_generator_tt}
    \tsor{L} = \tsor{\Xi} \cdot \Sigma_L \cdot \tsor{\Xi}^{+},
  \end{equation}
  where $\Sigma_L$ is a diagonal matrix with diagonal elements $\mu_i = \log(\lambda_i)/\stime$, and $\tsor{\Xi}^{+}$ is the pseudo-inverse of the Koopman eigentensor.
  When computing Eq.~\eqref{eq:koopman_generator_tt}, the contraction is performed as illustrated in Fig.~\ref{fig:proposed} to reduce memory consumption.

  The pseudo-inverse of the eigentensor $\tsor{\Xi}$ is computed as follows.
  The AMUSEt algorithm computes the Koopman eigentensor $\tsor{\Xi}$ as
  \begin{equation}
    \tsor{\Xi} = \tsor{U}_{X,r} \Sigma_{X,r} A,
  \end{equation}
  where $\tsor{U}_{X,r}$ is the left-orthonormalized TT format, $\Sigma_{X,r}$ is the diagonal matrix, and $A$ is the complex-valued matrix.
  Because the left-orthonormalized TT format satisfies an orthogonality relation analogous to that of an orthogonal matrix, the pseudo-inverse of $\tsor{\Xi}$ can be computed as
  \begin{equation}
    \begin{aligned}[b]
      \tsor{\Xi}^{+} &= \left\{\tsor{U}_{X,r} \Sigma_{X,r} A \right\}^{+} \\
      &= \left(\Sigma_{X,r} A\right)^+ \tsor{U}_{X,r}^{\ts} \\
      &= A^{+}\Sigma^{-1}_{X,r} \tsor{U}_{X,r}^{\ts} \\
      &= \left\{\tsor{U}_{X,r} \Sigma^{-1}_{X,r} \left\{ A^{+}\right\}^{\ts} \right\}^{\ts}.
    \end{aligned}
  \end{equation}
  The computation process is summarized in Algorithm~\ref{alg:proposed}.

  \begin{algorithm}[H]
  \caption{Proposed method for Koopman generator identification in the TT format}
  \label{alg:proposed}
  \begin{algorithmic}[1]
    \Require Snapshot matrices $X,Y \in \mathbb{R}^{\ndim \times \ndata}$
    \Ensure The Koopman generator in the TT format $\tsor{L}$

    \State Compute the Koopman eigentensor $\tsor{\Xi}$ and eigenvalues $\lambda_i$ via the AMUSEt algorithm.
    \State Compute the pseudo-inverse of the Koopman eigentensor as
      \begin{equation*}
        \tsor{\Xi}^{+} = \left\{\tsor{U}_{X,r} \Sigma^{-1}_{X,r} \left\{ A^{+}\right\}^{\ts} \right\}^{\ts}.
      \end{equation*}
    \State Compute the eigenvalues of $\tsor{L}$: $\mu_i = \log(\lambda_i)/\stime$.
    \State Compute the Koopman generator in the TT format as
    $\tsor{L} = \tsor{\Xi} \cdot \text{diag}(\mu_i) \cdot \tsor{\Xi}^{+}$.
  \end{algorithmic}
  \end{algorithm}

\section{Numerical experiments}
\label{sec:experiments}
In this section, we show the results of numerical experiments using the proposed method.

  \subsection{Lotka--Volterra system}
  To evaluate the accuracy of the proposed method, we first use a 4-dimensional Lotka--Volterra system \cite{lotka_1910_contribution,goel_1971_volterra} as a target system.
  The dynamics and the parameter are given by
  \begin{equation}
    \label{eq:lotka_volterra}
    \begin{aligned}
      \dot{x}_1 &= -0.12x_1 x_2 + 0.6x_1, \\
      \dot{x}_2 &= 0.12x_1x_2 - 0.14x_2x_3 + 0.4x_2, \\
      \dot{x}_3 &= 0.08x_2x_3 - 0.14x_3x_4 + 0.2x_3, \\
      \dot{x}_4 &= 0.06x_3x_4 - 0.42x_4.
    \end{aligned}
  \end{equation}
  We assume that the vector field is polynomial, so the dictionary is constructed using monomials of degree up to two in each variable; that is, the highest-degree term is $\prod_{i=1}^{4} x_i^2$, and the tensor is constructed as
  \begin{equation}
    \tsor{\psi}(\bm{x}) = \bigotimes_{i=1}^{4} \begin{bmatrix} 1 \\ x_i \\ x_i^2 \end{bmatrix}.
  \end{equation}
  In this setting, the Koopman matrix is of size $3^4 \times 3^4 = 81 \times 81$, which is easily computed.
  Because of this, we compare the vector field coefficients between the proposed method and the original matrix-logarithm-based method.

  \begin{figure}[b]
    \centering
    \includegraphics[keepaspectratio, scale=0.62]{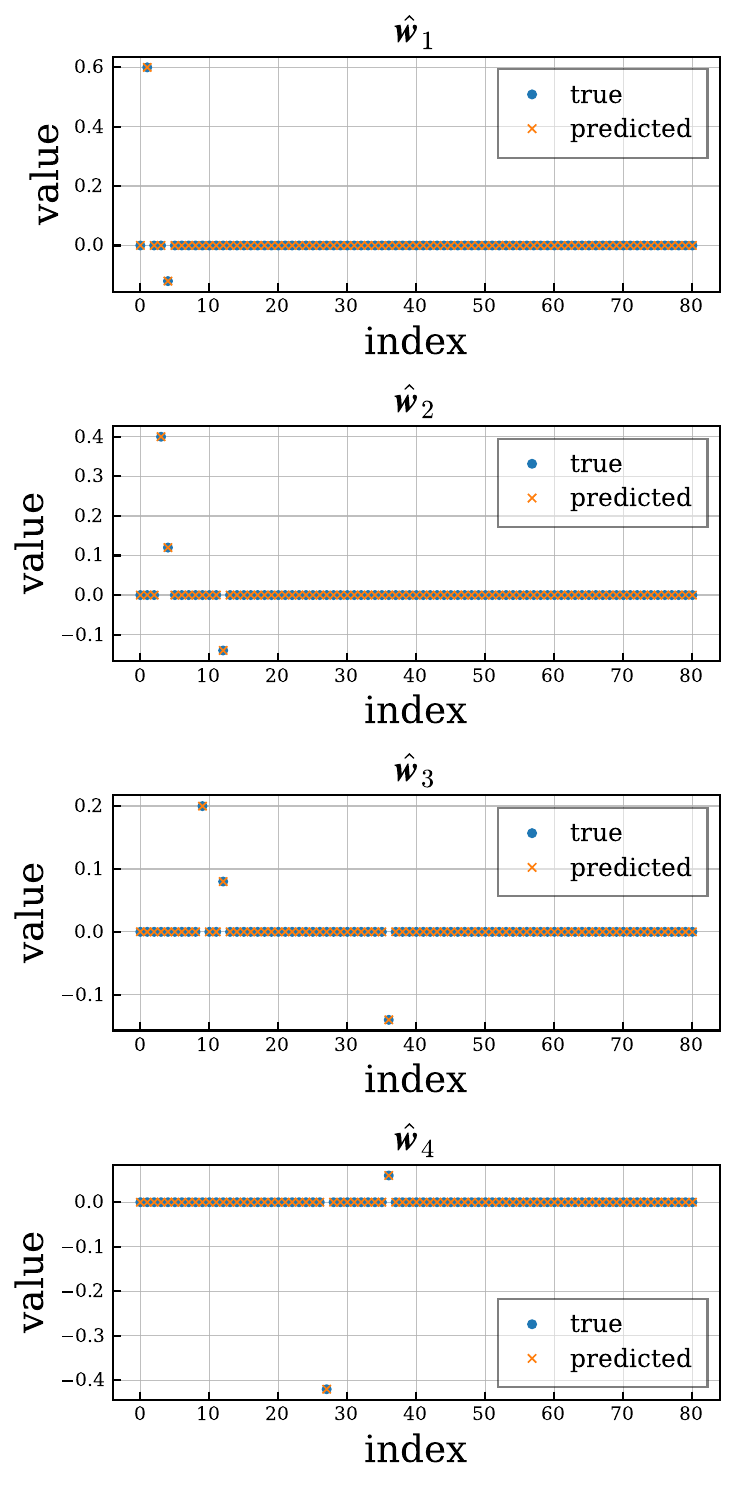}
    \caption{Vector field coefficients obtained for the Lotka--Volterra system.
The horizontal axis represents the indices of the coefficient vector, and the vertical axis represents the values of the coefficient vector.
The blue dots represent the true coefficients, and the orange dots represent the estimated coefficients.
All coefficients are nearly identical between the true and estimated values.}
    \label{fig:coeffs}
  \end{figure}

  We employ \verb|scipy.integrate.solve_ivp| function in Python with the explicit Runge--Kutta method of order 5(4) (RK45) to solve the ordinary differential equations shown in Eq.~\eqref{eq:lotka_volterra} to generate the trajectories.
  As a parameter setting, we use absolute tolerance $= 10^{-6}$ and relative tolerance $= 10^{-3}$.
  Under the parameter setting, we generate three trajectories using three different initial conditions: $x_1=\cdots=x_4=3.0$, $x_1=\cdots=x_4=5.0$, and $x_1=\cdots=x_4=8.0$.
  From each trajectory, we sample 5,000 data pairs with a sampling interval of $\stime = 0.1$.
  We use Algorithm~\ref{alg:proposed} to identify the Koopman generator from the snapshot pairs.
  For the AMUSEt algorithm, the truncation parameter is set to $\epsilon = 1 - 10^{-6}$.

  Figure~\ref{fig:coeffs} shows the estimates $\hat{\bm w}_k$ of the coefficients $\bm w_k$.
  The top panel of the figure shows the estimated vector field coefficients for the first dimension $\hat{\bm{w}}_1$.
  For example, $\hat{w}_{1,1}$ (index $= 1$) corresponds to the term $x_1$, and $\hat{w}_{1,5}$ (index $= 5$) corresponds to the term $x_1x_2$; these two terms appear in the true vector field defined in Eq.~\eqref{eq:lotka_volterra}.
  The proposed method accurately estimates the coefficients, and the estimated values are nearly identical to the true values.
  To quantitatively evaluate the accuracy, we compute the root-mean-square error (RMSE) defined as
  \begin{equation}
    \text{RMSE} = \sqrt{\frac{1}{\ndim \ndic} \sum_{k=1}^{\ndim} \sum_{i=1}^{\ndic} \left(w_{k,i} - \hat{w}_{k,i}\right)^2}
  \end{equation}
  and the normalized RMSE (NRMSE) defined as $\text{NRMSE} = \text{RMSE} / \overline{w}$,
  where $\overline{w}$ is the mean of the absolute values of the non-zero coefficients $|w_{k,i}|$.
  The resulting errors are $\text{RMSE} = 1.83 \times 10^{-5}$ and $\text{NRMSE} = 8.05 \times 10^{-5}$.
  The results show that the proposed method can accurately identify the coefficients of the vector field.

  \subsection{Higher-dimensional system}
  Next, we use a 10-dimensional Lorenz--96 system \cite{lorenz_1995_predictability} as a target system to show the effectiveness of the proposed method in high-dimensional systems.
  The dynamics are given by the following equations for $i = 1, \dots, 10$:
  \begin{equation}
    \begin{aligned}
      \dot{x}_i = (x_{i+1} - x_{i-2})x_{i-1} - x_{i} + 8.0,
    \end{aligned}
  \end{equation}
  where the indices are periodic, i.e., $x_{-1} = x_{9}, x_0 = x_{10}, \text{and } x_{11} = x_{1}$.
  We employ the same dictionary as in the previous experiment, consisting of monomials of degree up to two in each variable.
  In this setting, the Koopman matrix is of size $3^{10} \times 3^{10} = 59,049 \times 59,049$, which is difficult to compute as a matrix due to memory constraints.

  We generate one trajectory from the initial state $x_1 = 8.1$ and $x_2, \dots, x_{10} = 8.0$ by using \verb|scipy.integrate.solve_ivp| function same as the previous experiment.
  After discarding the first 100,000 data pairs for initial relaxation, we sample 20,000 data pairs with a sampling interval of $\stime = 0.01$.
  The sampling interval is set to be small because the system is highly chaotic.
  Algorithm~\ref{alg:proposed} is used to identify the Koopman generator from the snapshot pairs.
  For the AMUSEt algorithm, the truncation parameter is set to $\epsilon = 1 - 10^{-6}$.

  The number of vector field coefficients is $3^{10} = 59,049$, and it is difficult to check all the coefficients due to the computational cost of evaluating the values from the TT format.
  Therefore, we check only the coefficients corresponding to terms of total degree up to two, i.e., linear terms $x_i$ and quadratic terms $x_i x_j$.
  The resulting errors for these coefficients are $\text{RMSE} = 2.35 \times 10^{-6}$ and $\text{NRMSE} = 8.56 \times 10^{-7}$.
  Table~\ref{tab:result_l96_nonzero} shows the estimation errors for the terms $x_{i+1} x_{i-1}$, which appear in the true dynamics.
  Table~\ref{tab:result_l96_zero} shows the estimation errors for the terms $x_{i}x_{i+1}$ as examples of terms correctly identified as zero.
  The results show that the proposed method can accurately identify the coefficients of the vector field.

  Furthermore, the memory consumption (defined as the number of elements in the tensor) of the Koopman generator $\tsor{L}$ for the proposed method is $1.39 \times 10^7$.
  In contrast, the memory consumption for the original logarithm method is expected as $59,049^2 \approx 3.49 \times 10^{9}$.
  This result shows that the proposed method significantly reduces the memory consumption compared to the original logarithm method, especially in high-dimensional systems.

  In this experiment, the true vector field is known, which allows us to compute the admissible upper limit of the sampling interval using Eq.~\eqref{eq:ts_limit}.
  This gives the condition $\stime < 0.07$, which is satisfied by the sampling interval used in the experiment.
  To examine the effect of violating this condition, we also test $\stime = 0.1$ under the same setting.
  The resulting RMSE is 0.94, indicating that the branch ambiguity of the logarithm leads to significant errors in the estimated coefficients when the sampling interval is not sufficiently small.

  \begin{table}[t]
    \centering
    \caption{Absolute errors in the estimated coefficients of the $x_{i+1} x_{i-1}$ terms in the Lorenz--96 system.
The absolute error for each coefficient is defined as the absolute value of the difference between the true and estimated values, i.e., $|\hat{w}_{k,i} - 1|$.}
    \begin{tabular}{cc}
      \hline
      Term & Absolute error \\ \hline \hline
       $x_{2}x_{10}$ & $3.67 \times 10^{-8}$ \\ \hline
       $x_{3}x_{1}$  & $5.18 \times 10^{-6}$ \\ \hline
       $x_{4}x_{2}$  & $4.69 \times 10^{-8}$ \\ \hline
       $x_{5}x_{3}$  & $6.75 \times 10^{-7}$ \\ \hline
       $x_{6}x_{4}$  & $1.56 \times 10^{-7}$ \\ \hline
       $x_{7}x_{5}$  & $5.78 \times 10^{-8}$ \\ \hline
       $x_{8}x_{6}$  & $7.43 \times 10^{-7}$ \\ \hline
       $x_{9}x_{7}$  & $4.02 \times 10^{-7}$ \\ \hline
       $x_{10}x_{8}$ & $2.20 \times 10^{-7}$ \\ \hline
       $x_{1}x_{9}$  & $1.16 \times 10^{-6}$ \\ \hline
    \end{tabular}
    \label{tab:result_l96_nonzero}
  \end{table}

\section{Conclusion}
\label{sec:conclusion}
In this paper, we proposed a method to identify the Koopman generator in the TT format from data by using the operator logarithm.
The proposed method uses the AMUSEt algorithm to compute the Koopman eigenfunctions in the TT format, and then derives the Koopman generator by taking the logarithm of the eigenvalues.
We demonstrated the effectiveness of the proposed method through numerical experiments on a 4-dimensional Lotka--Volterra system, and a 10-dimensional Lorenz--96 system.
The results showed that the proposed method can accurately identify the coefficients of the vector field while significantly reducing the memory consumption compared to the original logarithm method, especially in higher-dimensional systems.

This study provides an initial framework for identifying the Koopman generator in the TT format; however, several open problems remain.
Further work includes investigating the robustness of the proposed method to noise in the data and incorporating sparsity-promoting regularization (e.g., LASSO) to obtain sparse coefficient estimates.
When the generator is obtained via the operator logarithm, it remains unclear how to enforce sparsity in the estimated coefficients in a principled manner.
In addition, the current implementation computes a truncated set of Koopman eigenvalues through matrix eigendecomposition, which becomes computationally prohibitive for large-scale systems.
An important direction for future research is to develop scalable algorithms that compute Koopman eigenvalues efficiently within the TT format.

  \begin{table}[t]
    \centering
    \caption{Absolute errors in the estimated coefficients of the $x_i x_{i+1}$ terms, which are not present in the Lorenz--96 system.
The absolute error for each coefficient equals the absolute value of the estimate, since the true value is zero.}
    \begin{tabular}{cc}
      \hline
      Term & Absolute error \\ \hline \hline
      $x_{1}x_{2}$  & $9.46 \times 10^{-7}$ \\ \hline
      $x_{2}x_{3}$  & $4.22 \times 10^{-6}$ \\ \hline
      $x_{3}x_{4}$  & $1.62 \times 10^{-6}$ \\ \hline
      $x_{4}x_{5}$  & $9.29 \times 10^{-7}$ \\ \hline
      $x_{5}x_{6}$  & $1.19 \times 10^{-7}$ \\ \hline
      $x_{6}x_{7}$  & $4.37 \times 10^{-7}$ \\ \hline
      $x_{7}x_{8}$  & $2.16 \times 10^{-7}$ \\ \hline
      $x_{8}x_{9}$  & $1.18 \times 10^{-7}$ \\ \hline
      $x_{9}x_{10}$ & $2.25 \times 10^{-7}$ \\ \hline
      $x_{10}x_{1}$ & $5.25 \times 10^{-7}$ \\ \hline
    \end{tabular}
    \label{tab:result_l96_zero}
  \end{table}

\appendix
\section{Koopman eigenvectors in the TT format}
\label{sec:appendix}
In this section, we describe how to compute Koopman eigenvalues and eigenvectors in the TT format using the AMUSEt algorithm \cite{nuske_2021_tensorbased}.
The approach first computes the global SVD \cite{klus_2018_tensorbased,gelss_2019_multidimensional} of the transformed data tensors $\tsor{\Psi}(X)$ and $\tsor{\Psi}(Y)$ in the TT format.
From these decompositions, we construct a reduced Koopman matrix $M$ and then solve an eigenvalue problem for $M$ to obtain the Koopman eigenvalues and the corresponding eigenvectors in the TT format.

The reduced matrix $M$ can be interpreted as a projection of the Koopman operator onto the subspace spanned by the left singular vectors of $\tsor{\Psi}(X)$.
Accordingly, $M \in \mathbb{R}^{r \times r}$, where $r$ denotes the truncation rank (i.e., the number of retained singular values).
The procedure for the global SVD is summarized in Algorithm~\ref{alg:globalsvd} and is otherwise identical to \cite{nuske_2021_tensorbased}.
When computing the truncated SVD, we use a truncation parameter $\epsilon$ and determine the threshold based on the cumulative sum of singular values; specifically, we choose the smallest $j$ satisfying
\begin{equation}
  \frac{\Sigma_{j} \sigma_{j}}{\Sigma_{i} \sigma_{i}} < \epsilon.
\end{equation}

\begin{figure}[t]
  \centering
  \includegraphics[keepaspectratio, scale=0.45]{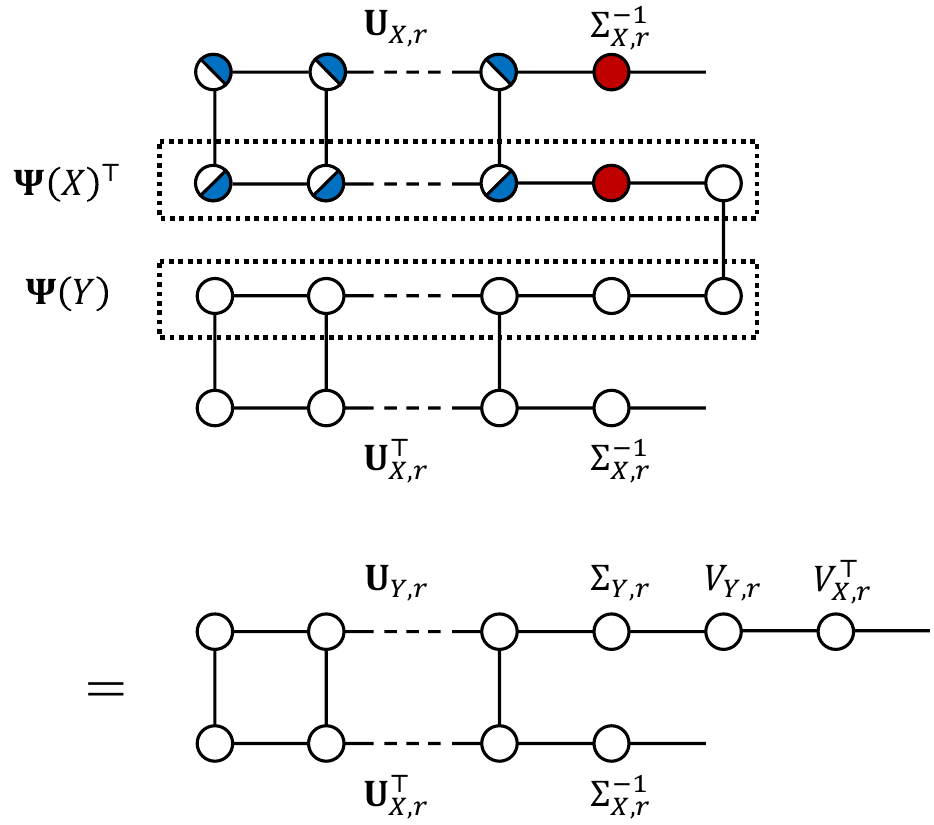}
  \caption{Graphical representation of the reduced-matrix computation.
The tensors $\tsor{U}_{X,r}$ and $\tsor{U}_{Y,r}$ are left-orthonormalized so that the contractions can be performed efficiently.
Since we define $\tsor{\Psi}(Y) = K \tsor{\Psi}(X)$, the tensor contraction sequence differs from that in the original algorithm.}
  \label{fig:reduced_matrix}
\end{figure}

Using the global SVD, we obtain the decompositions $\tsor{\Psi}(X) = \tsor{U}_{X,r} \Sigma_{X,r} V^{\ts}_{X,r}$ and $\tsor{\Psi}(Y) = \tsor{U}_{Y,r} \Sigma_{Y,r} V^{\ts}_{Y,r}$, where $\tsor{U}_{X,r}$ and $\tsor{U}_{Y,r}$ are orthogonal tensors in the TT format, $\Sigma_{X,r}$ and $\Sigma_{Y,r}$ are diagonal matrices, and $V_{X,r}$ and $V_{Y,r}$ are orthogonal matrices.
The reduced matrix $M$ is then computed as
\begin{equation}
  \label{eq:reduced_matrix}
  M = \Sigma_{X,r}^{-1} \tsor{U}^{\ts}_{X,r} \tsor{U}_{Y,r} \Sigma_{Y,r} V_{Y,r} V_{X,r}^{\ts}.
\end{equation}
In this paper, some tensor contractions in Eq.~\eqref{eq:reduced_matrix} differ from those in \cite{nuske_2021_tensorbased} due to our convention of defining the dictionary as a column vector.
The detailed computation of $M$ is summarized in Algorithm~\ref{alg:amuset}, and an overview of the contraction sequence is illustrated in Fig.~\ref{fig:reduced_matrix}.

\begin{algorithm}[H]
\caption{Global SVD \cite{klus_2018_tensorbased,gelss_2019_multidimensional}}
\label{alg:globalsvd}
\begin{algorithmic}[1]
  \Require Transformed data tensor $\tsor{\Psi}(X)$ in the TT format
  \Ensure Global SVD of $\tsor{\Psi}(X)$ in the form $\tsor{\Psi}(X) = \tsor{U}_{X,r} \Sigma_{X,r} V^{\ts}_{X,r}$

  \For{$k=1, \dots, d-1$}
  \State Reshape $\tsor{\Psi}(X)^{(k)}$ to a matrix $T \in \mathbb{R}^{r_{k-1} n_k \times r_k}$.
  \State Compute the truncated SVD of $T$ as $T \approx U \Sigma V^{\ts}$ with the truncation parameter $\epsilon$.
  \State Set $\tsor{\Psi}(X)^{(k)}$ to a reshaped version of $U$.
  \State Set $\tsor{\Psi}(X)^{(k+1)}$ to a reshaped version of $\Sigma V^{\ts} \tsor{\Psi}(X)^{(k+1)}$.
  \EndFor
  \State Reshape $\tsor{\Psi}(X)^{(d)}$ to a matrix $T \in \mathbb{R}^{r_{d-1} n_d \times r_d}$.
  \State Compute the truncated SVD of $T$ as $T \approx U \Sigma V^{\ts}$ with the truncation parameter $\epsilon$.
  \State Set $\tsor{\Psi}(X)^{(d)}$ to a reshaped version of $U$.
  \State Set $\tsor{\Psi}(X)^{(d+1)}$ to a reshaped version of $V^{\ts}$.
  \State Define $\tsor{U}_{X,r} = \matbb{\tsor{T}^{(1)}} \otimes \cdots \otimes \matbb{\tsor{T}^{(d)}}$, $\Sigma_{X,r} = \Sigma$, and $V_{X,r} = V$.
\end{algorithmic}
\end{algorithm}

\begin{algorithm}[H]
\caption{The AMUSEt algorithm \cite{nuske_2021_tensorbased}}
\label{alg:amuset}
\begin{algorithmic}[1]
  \Require Snapshot matrices $X, Y \in \mathbb{R}^{\ndim \times \ndata}$
  \Ensure TT approximation of Koopman eigentensor $\tsor{\Xi}$ and eigenvalues $\lambda_i$ (of $K$)

  \State Compute global SVDs of $\tsor{\Psi}(X)$ and $\tsor{\Psi}(Y)$, i.e.,
  $\tsor{\Psi}(X) = \tsor{U}_{X,r} \Sigma_{X,r} V^{\ts}_{X,r}$ and $\tsor{\Psi}(Y) = \tsor{U}_{Y,r} \Sigma_{Y,r} V^{\ts}_{Y,r}$.
  \State Compute the reduced matrix $M = \Sigma_{X,r}^{-1} \tsor{U}_{X,r}^{\ts} \tsor{U}_{Y,r} \Sigma_{Y,r} V_{Y,r} V_{X,r}^{\ts}$.
  \State Compute eigenvalues $\lambda_i$ of $M$ and eigenvectors $\bm v_i$.
  \State Express the eigentensor with respect to the original basis as
  $\tsor{\Xi} = \tsor{U}_{X,r} \Sigma_{X,r} \left[\bm v_1 \cdots \bm v_r\right]$.
\end{algorithmic}
\end{algorithm}

\begin{acknowledgments}
This work was supported by JST FOREST Program (Grant Number JPMJFR216K, Japan).
\end{acknowledgments}

\end{document}